\documentclass[10pt,twocolumn,letterpaper]{article}

\usepackage{wacv}
\usepackage{times}
\usepackage{epsfig}
\usepackage{graphicx, caption, subcaption}
\usepackage{amsmath}
\usepackage{amssymb}
\usepackage{gensymb}
\usepackage{rotating}
\usepackage{multirow}
\usepackage{makecell}
\usepackage[numbers,sort]{natbib}
\usepackage{comment}
\usepackage{adjustbox}
\usepackage{enumitem}

\newsavebox\CBox
\def\textBF#1{\sbox\CBox{#1}\resizebox{\wd\CBox}{\ht\CBox}{\textbf{#1}}}
\usepackage[draft, pagebackref=true,breaklinks=true,letterpaper=true,colorlinks=false,bookmarks=false]{hyperref}

\wacvfinalcopy % *** Uncomment this line for the final submission

\def\wacvPaperID{437} % *** Enter the wacv Paper ID here

\ifwacvfinal\fi
\begin{document}
%%%%%%%%% TITLE
\title{Learning from THEODORE: A Synthetic Omnidirectional Top-View Indoor Dataset for Deep Transfer Learning}

\author{Tobias Scheck\thanks{Authors contributed equally} \space , Roman Seidel$^*$\thinspace , Gangolf Hirtz\\
Chemnitz University of Technology\\
Faculty of Electrical Engineering and Information Technology\\
09126 Chemnitz, Germany\\
{\tt\small {tobias.scheck,roman.seidel,g.hirtz}@etit.tu-chemnitz.de}
}

\maketitle
\thispagestyle{empty}

\begin{abstract}

Recent work about synthetic indoor datasets from perspective views has shown significant improvements of object detection results with Convolutional Neural Networks (CNNs).
In this paper, we introduce THEODORE: a novel, large-scale indoor dataset containing 100,000 high-resolution diversified fisheye images with 16 classes.
To this end, we create 3D virtual environments of living rooms, different human characters and interior textures.
Beside capturing fisheye images from virtual environments we create annotations for semantic segmentation, instance masks and bounding boxes for object detection.
We compare our synthetic dataset to state of the art real-world datasets for omnidirectional images.
Based on MS COCO weights, we show that our dataset is well suited for fine-tuning CNNs for object detection and semantic segmentation.
Through a high generalization of our models by means of image synthesis and domain randomization we reach a AP up to 0.90 for class person on our own annotated fisheye evaluation suite (FES).
Additionally, the evaluation of six classes was done through object detection and semantic segmentation on FES. 
The segmentation task on FES leads to 0.36 mIoU on all classes and to a mAP of 0.61 for the object detection.

\end{abstract}

\section{Introduction}

% How to write a good introduction? from: https://www.ncbi.nlm.nih.gov/pmc/articles/PMC4548565/

% Abbreviations should be given following their explanations in the ‘Introduction’ section (their explanations in the summary does not count)

% Simple present tense should be used.

% References should be selected from updated publication with a higher impact factor, and prestigous source books.

% Avoid mysterious, and confounding expressions, construct clear sentences aiming at problematic issues, and their solutions.

% The sentences should be attractive, tempting, and comjprehensible.

% Firstly general, then subject-specific information should be given. Finally our aim should be clearly explained.

% Part 1: Preparation of the background

Synthetic images and labels from modeled 3D indoor scenes has been an increasing research field in computer vision in the last few years.
In contrast to manually labeled indoor front-view perspective images for action recognition \cite{978-3-319-10602-1_48, schuldt2004recognizing, BMVC201552} that are widely explored, image data from top-view indoor scenes of omnidirectional images are rarely available.
Invariance against the perspective of objects, e.g. missing images from top-view scenes makes common datasets not adaptable to computer vision tasks on omnidirectional cameras.
The most widely used projection of fisheye images is the equirectangular camera model, where all image points are mapped to the inside of a lower half sphere through elevation and azimuth.
This projection formulates the distortion of omnidirectional images and leads to a high variation of the shape of objects depending on their location in the image.

% Part 2: Discussion of the basic references related to the main topic

In this paper we introduce \textit{THEODORE - a synTHEtic tOp-view inDoOR scEnes} dataset that contains diversified rendered fisheye images of indoor environments with instance segmentation masks and bounding boxes.
The indoor world was created with the game engine Unity3D and rendered images were captured with a camera that follows the omnidirectional projection.
To bridge the gap between real-world and synthetic images we perform domain randomization with different rooms, persons, objects and camera positions.
With THEODORE we release a dataset that improves the accuracy of state-of-the-art CNNs on omnidirectional images in indoor environments.
A few application fields are navigation of autonomous systems through visual odometry, personal security in public transportation services or in virtual reality.
We expect a strong growth of the research field on omnidirectional images in computer vision.\\

\noindent Our contribution is twofold:

\begin{itemize}[topsep=3pt]
    \setlength\itemsep{-0.2em}
    \item Generating THEODORE: a dataset with diversified omnidirectional images and labels for indoor scenes
    \item Improvement of accuracy of state-of-the-art CNNs for object detection in omnidirectional images
\end{itemize}
% Part 3: Indication of the purpose (Angabe des Verwendungszweckes, Applikationsfelder) and structure of the paper

The paper is structured as follows.
Following this introduction in chapter \ref{sec:relatedwork} we treat related works to synthetic image data and one- and two-stage object detection.
In chapter \ref{sec:dataset} we describe the data generation process and its properties.
In chapter \ref{sec:approach} we describe the behaviour of state of the art single- and two-stage object detectors in terms of THEODORE.
The evaluation on publicly available databases for omnidirectional images is shown in Chapter \ref{sec:evaluation}.
We summarize our results and give future research directions in Chapter \ref{sec:conclusion}.
Our dataset can be found at \url{https://www.tu-chemnitz.de/etit/dst/forschung/comp_vision/theodore}.

\begin{figure*}[t]
	\begin{center}
		\includegraphics[width=0.9\linewidth]{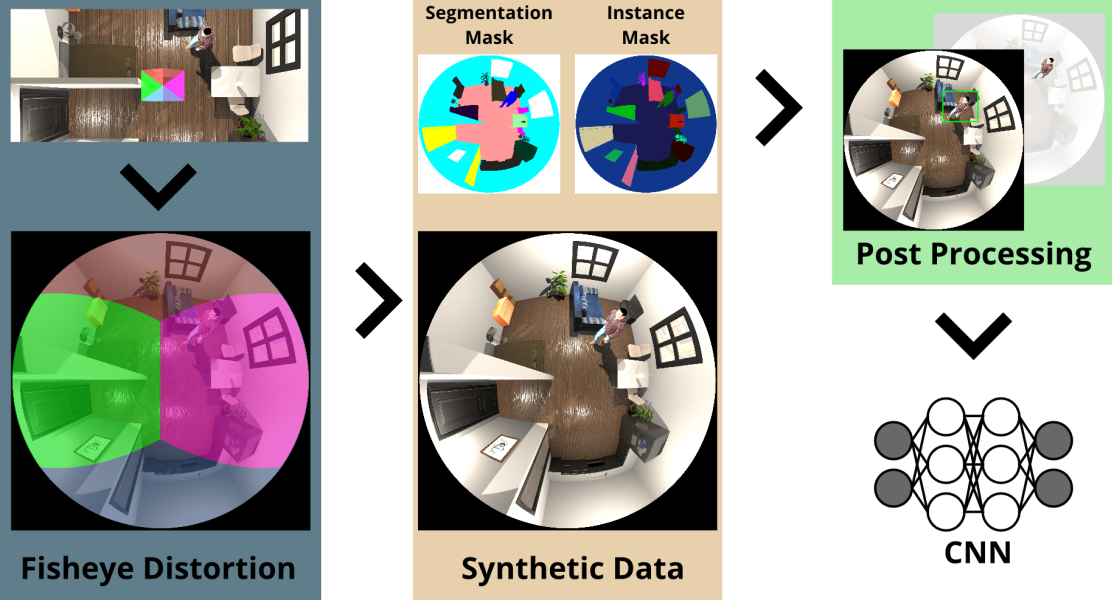}
    \end{center}
    \caption{Pipeline showing the image generation for THEODORE. Starting with four cameras pointing in top, left, right and bottom direction with a field of view of 90 degrees per camera, a fisheye distorted image is generated. In addition to the rendered RGB image, instance and segmentation masks are created, distorted and saved as training data. In post-processing, bounding boxes are extracted and converted into common dataset formats such as TFRecord.}
    \label{fig:pipeline}
\end{figure*}

\section{Related Work}
\label{sec:relatedwork}
Synthetic data for omnidirectional images isn't well explored and data for training CNNs are sparsely available. In this section the most relevant indoor datasets and CNN architectures for object detection are introduced.
\newline\noindent\textbf{Synthetic Data}
Synthetic data of persons in perspective views was widely studied \cite{varol2017learning} for tasks like object detection, segmentation or human pose estimation \cite{ionescu2011latent, ionescu2014human3}.
For the analysis of multi-object tracking Gaidon \textit{et al.} created the Virtual KITTI dataset \cite{gaidon2016virtual}, including different environment conditions, camera position and instance-level segmentation ground truth.
A couple of 3D model repositories for indoor scenes \cite{handa2016understanding, InteriorNet18} in perspective views with focus on depth, physical based rendering and volumetric ground truth, namely the SUNCG dataset \cite{song2016ssc} and the Matterport dataset \cite{Matterport3D}, were released.
Based on these datasets the work of \cite{song2016im2pano3d} generates a RGB-D panorama dataset for different camera configurations, but without different camera models and top-view images.
While multisensory models for goal-directed navigation in complex indoor environments from ego-perspective MINOS \cite{savva2017minos} was published, extensive research in terms of semantic descriptions, acoustics and multi agent support from 3D visual renderings led to HoME \cite{brodeur2017home}.
With the goal to create houshold activities in virtual homes the work of \cite{virtualhome2018} delivers instance and semantic label annotation, depth, pose and optical flow.
The novelty of this approach is the formulation of the automatic generation of program episodes from text and creatable avatar videos.
Our approach differs from this work in terms of camera geometry, domain randomization and viewing angle.
House3D \cite{wu2018building} provides 3D scenes of visually realistic houses that are equipped  with  a  diverse  set  of  fully  labeled  3D  objects and textures based on the  SUNCG  dataset including RGB images, depth, segmentation masks and top-down 2D map views.
In terms of selecting the viewing angle of indoor scenes automatically, the work of \cite{genova2017learning} uses per class statistics to find the best viewing angle for semantic segmentation.

Existing datasets of omnidirectional image data (\cite{piropo2016, demiroz2012feature, figueira2014hda+, eichenseer2016data, cinaroglu2014direct}) have low in-class variance, missing ground truth labels or contain less variations of scenes \cite{Liciotti2017reid}.

\noindent\textbf{Object Detection} One-stage object detectors \cite{redmon2018yolov3, 10.1007/978-3-319-46448-0_2} that treat object detection as a simple regression task learn class probabilities and bounding box coordinates.
Two-stage detectors such as \cite{NIPS2015_5638} and \cite{NIPS2016_6465} generate regions of interest (ROIs) by a Region Proposal Network in the first and forward these ROIs to the object classification and bounding box regression pipeline.\\
Object detection in distorted fisheye images is not widely explored.
The authors of \cite{Coors2018ECCV} and \cite{Cohen2018ICLR} adapt the network architecture of CNNs to spherical representations of the regular convolution operations.
However, \cite{Coors2018ECCV} wraps the sampled locations of convolutional filters to the sphere and effectively reverses the distortions of the omnidirectional camera model.
\cite{Cohen2018ICLR} avoids translational weight sharing and creates building blocks that satisfy a generalized Fourier theorem, to detect patterns independently from their location on the sphere.

Current frameworks with AI agents (\cite{ai2thor, wu2018building}) concentrate on embodied question answering or navigation (PointGoal, ObjectGoal and RoomGoal).
Images and corresponding labels (segmentation masks, surface normals, object IDs, depth) can be created, but are missing for omnidirectional camera model.
In contrast to our work, viewing angle of AI agent frameworks is front-view from an ego-perspective.

Taylor et al. presents with a virtual worlds environment the possibility to create foreground masks, bounding boxes and target centroids in top-view omnidirectional images \cite{taylor2007ovvv}.

\begin{figure*}[t]
	\begin{center}
		\includegraphics[width=0.9\linewidth, trim=0 3 0 3, clip]{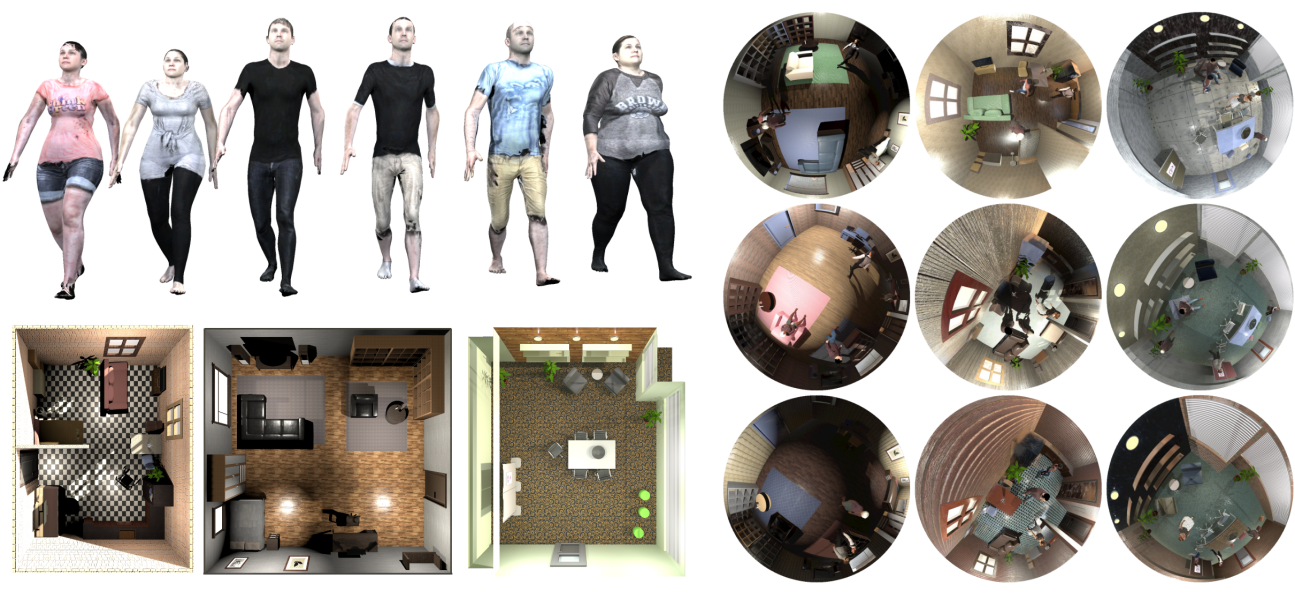}
    \end{center}
    \caption{Overview of random characters and room floor plans used in THEODORE.
    Additionally, some sample images with the applied domain randomization are depicted.
    For each room three random textured scenes with random camera positions are selected.
    \label{fig:overview}}
\end{figure*}

\section{THEODORE Dataset}
\label{sec:dataset}

In this section all relevant steps for the generation of our synthetic dataset THEODORE are presented.
We show the properties of the dataset in terms of distribution and variations of viewing angle.

\subsection{Data Generation}
\noindent\textbf{Game Engine}
An advantage of the usage of a game engine, compared to rendering software like Blender,
is the opportunity to generate data in a less time-consuming manner.
In this work we are generating synthetic data using the game engine Unity3D.
To configure the walking path of the characters in the virtual environment,
Unity3D provides a NavMesh component that allows avoiding obstacles by approximating the walkable areas.

The generated virtual environments consist of indoor scenes, where typical objects like tables, sofas and chairs are placed at fixed positions.
Human 3D characters are generated using the Skinned Multi-Person Linear Model (SMPL) \cite{SMPL:2015}.
SMPL is a model of the human body with focus of realism based on thousands of 3D body scans.
Human characters are able to move randomly in the area determined to be valid by NavMesh.
Each character moves from a random start position to a randomly selected object position as destination.

We need to capture the whole scene from an omnidirectional camera placed on the ceiling of the room.
However, Unity3D only provides a camera model for perspective and orthographic projection.
This limitation can be overcome by combining four perspective cameras in order to generate an omnidirectional image as described in the following.

\noindent\textbf{Fisheye Projection}
Real omnidirectional images can be obtained by using fisheye lenses which results in a barrel distortion.
Inside Unity3D fisheye images can be generated using the approach described by Bourke \textit{et al.\ }\cite{david_bourke_blender_2010, bourke_idome:_2009}.

This method is based on a modified cube map rendering (see \autoref{fig:pipeline}) using 4 of the cube faces to form a fisheye distorted image.
Each face is the result of a rendered image captured by a camera with a field of view of 90\,\degree.
As shown in \autoref{fig:overview}, the final result is created by warping and combining these images on four meshes,
whose texture coordinates model a fisheye projection.
Afterwards the generated and distorted fisheye image is captured with an orthographic camera and rendered to the display.

For THEODORE we are using a resolution of $1024 \times 1024$ pixels which allows us,
in combination with a native plugin, to reach an output of 15 FPS on an Intel i7-7700 and a Nvidia GTX 1080.
The native plugin allows us to manage the transfer of the textures from GPU to the CPU memory in a faster way than the conventional methods available in Unity3D.

\noindent\textbf{Image Synthesis}
In addition to a rendered image we are generating segmentation- and instance masks.
This is done by cloning the virtual omnidirectional camera setup and replacing the assigned shaders of each object with a unique color shader.
In the case of segmentation, the colors are assigned according to the object classes.
For instance masks the color is selected based on the unique object ID.
With these modifications, the approach presented in \cite{UnityTec73:online} fits the previously described fisheye projection.
In this case the shader replacement is performed for all four perspective cameras before generating the final segmentation and instance masks.

\noindent\textbf{Domain Randomization}
An approach for bridging the gap between synthetic and real images is domain randomization \cite{8202133,8575297} that we also apply in our implementation.
Every room changes after 25 seconds, which we call a level change.
With each level change a new room is selected and object textures are randomly replaced.
Furthermore, human characters are generated with random parameters (like height or weight) and textures, using the texture set from \cite{varol2017learning}.
However, the replacement occurs inside a predefined texture set (e.g. wood, concrete, cotton, etc.), to prevent inappropriate texture assignments.
Additionally the camera position is changed over time in order to create different points of view.
The trajectory of the camera follows a Lissajous curve. %: $A=4$, $B=6$, $a=5$, $b=3$ and $\delta = \frac{\pi}{2}$.
Light sources are defined as point lights with a fixed range and intensity.
The number of enabled light sources in each room is selected to ensure a well illuminated scene.
To create different lighting situations with each level change some randomly selected light sources are disabled,
however with the restriction that at least one light source remains active.

\noindent\textbf{Post Processing}
The final image, the segmentation and instance masks are combined in order to extract the necessary bounding boxes for the CNN training.
By segmenting per color on the instance mask, a binary mask for each object is generated.
Then these masks are applied on the segmentation mask to identify each object with its corresponding label and the bounding box coordinates $x_{min}$, $x_{max}$, $y_{min}$ and $y_{max}$ are calculated.
Finally the fisheye images together with the extracted bounding boxes and their corresponding labels are used to perform a conversion into common dataset formats (e.g. TFRecords \cite{8099834} or PASCAL VOC2012 \cite{everingham_pascal_nodate}).

\begin{figure}[t]
    \begin{center}
        \begin{subfigure}[t]{0.47\textwidth}
            \includegraphics[width=1\linewidth, trim=0 10 0 5, clip]{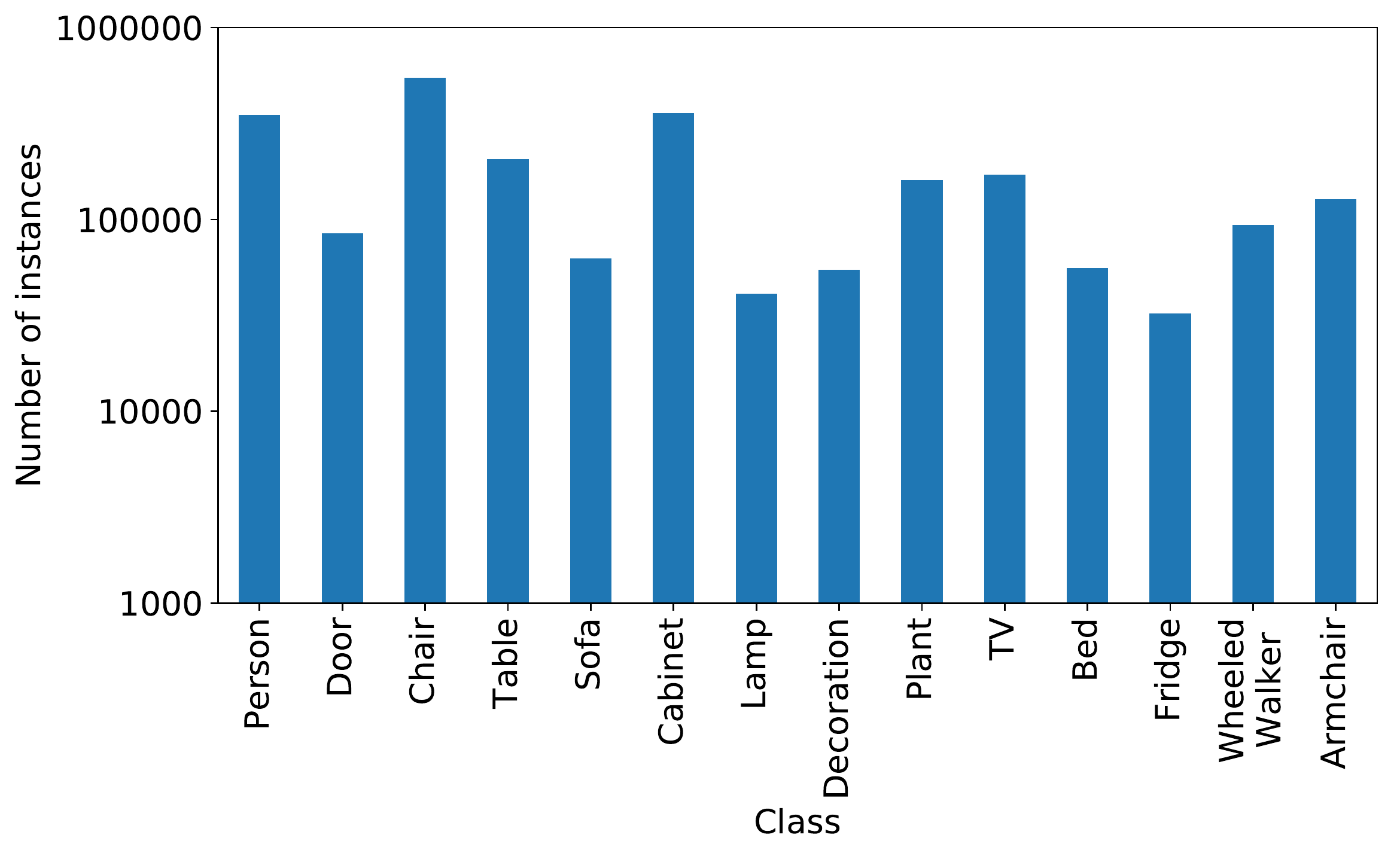}
            \caption{Instances per class\label{fig:stats:instances_per_class}}
        \end{subfigure}
        \begin{subfigure}[t]{0.47\textwidth}
            \includegraphics[width=1\linewidth, trim=0 20 0 30, clip]{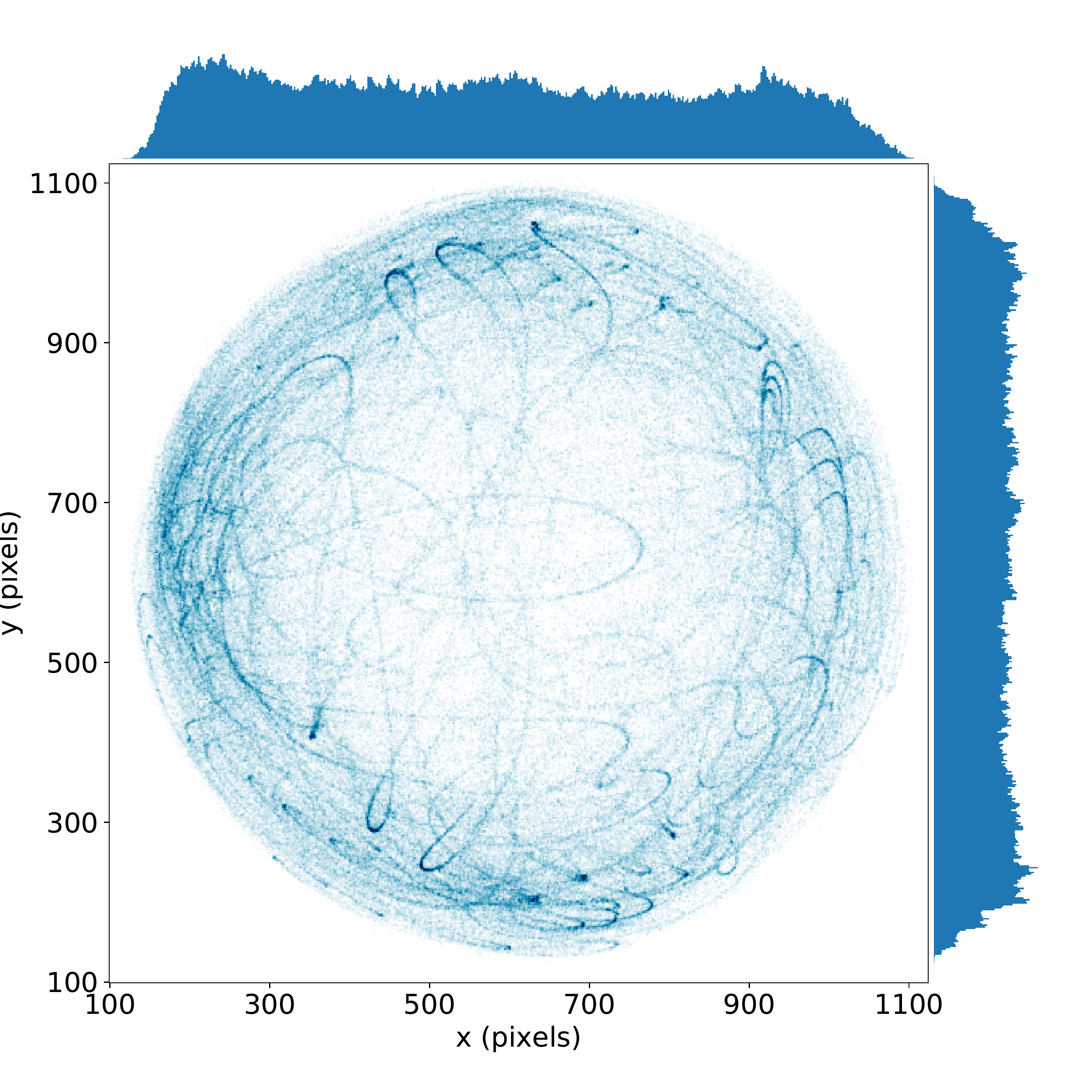}
            \caption{Distribution of centroid location of a person\label{fig:stats:centroid}}
        \end{subfigure}
    \end{center}
    \caption{
        In Figure \ref{fig:stats:instances_per_class} we show the number of annotations per class.
        In Figure \ref{fig:stats:centroid} the distribution of centroid location of a person over all images is illustrated.
        See text for details.}
    \label{fig:stats}
\end{figure}

\subsection{Dataset Analysis}
The creation pipeline of our synthetic omnidirectional data is visualized in \autoref{fig:pipeline}.
Apart from the final image we extract the segmentation and label mask from the rendering process.
Based on these masks we are able to select specific objects to calculate the bounding box.
For THEODORE we have exported 100k images and bounding boxes for the classes person, chair, table, armchair, wheeled walker, tv generated.
The dataset contains different rooms with randomly selected textures, as described in \autoref{sec:dataset}.
For this approach we downloaded and categorized 120 textures\footnote{https://www.cc0textures.com},
so that each textured 3D object can theoretically choose one of them. We recorded the scene with $8$ frames per second.
In combination with a level change parameter of $25$ seconds and the texture randomization, the dataset contains about $500$ various textured indoor scenes.

An example for three randomized men and women with varying body shape and height, wearing different clothes and additional attributes of our simulation is depicted in \autoref{fig:overview}.
The amount of instances per class is visualized in \autoref{fig:stats}(\subref{fig:stats:instances_per_class}) and the statistical distribution from the center point of persons bounding box in \autoref{fig:stats}(\subref{fig:stats:centroid}).
Through camera movement and random selection of destinations for a person we ensure well distributed positions over all fisheye images.

\section{Approach}
\label{sec:approach}
In this section we describe the functionality of three meta-architectures of CNNs for object detection and two semantic segmentation networks and show the corresponding training setup.
We train the architectures with our synthetic data using an open source framework for object detection \cite{8099834} and a own implementation for the segmentation task.
For object detection task we choose one- and two-stage object detectors and for segmentation we use pixel-wise classifiers as following described:
\\\\
\textbf{Faster R-CNN}
The Faster R-CNN architecture uses two stages for the detection.
The first stage, the region proposal network (RPN), is used to predict and extract box proposals.
For this stage a feature extractor is used to extract the features of an image at various intermediate maps.
In the second stage proposed boxes are cropped from an intermediate feature map and fed to the remainder of the feature extractor to refine and predict classes of the box proposals.
As feature extractor we use ResNet50 \cite{7780459}.
\\\textbf{R-FCN}
R-FCN is similar to R-CNN.
The difference is in the cropping approach.
An R-FCN only crops the result of the last layer while a Faster R-CNN crops the features from layers where the region proposal is predicted.
This reduces the pro-region computation because cropping happens only at the end of the network and results in a faster run time.
Here, ResNet101 \cite{7780459} is used as feature extractor.
\\\textbf{Single Shot Detector}
The Single Shot Detector (SSD) uses a single feed-forward convolutional network to predict box anchors and classes directly without using a second stage per-proposal classification.
The final detections are the results of a non-maximum suppression step applied to the prior predictions.
For our approach we use the feature pyramid network (FPN) \cite{8099589} implementation of ResNet50.
\\\textbf{SegNet}
SegNet \cite{badrinarayanan2017segnet} is a CNN architecture for semantic pixel-wise segmentation.
It consists of an encoder network which topology is identical to VGG-16 network \cite{simonyan2014very}.
However, fully connected layers were removed to improve the size of the network and the training process.
The decoder network of SegNet restores the gradually reduced spatial dimension from the encoder network.
To realise this, SegNet uses pooling indices of the corresponding encoder for a non-linear upsampling.
\\\textbf{PSPNet}
Pyramid Scene Parsing Network (PSPNet) \cite{zhao2017pyramid} is a scene parsing framework that uses a pyramid pooling module to aggregate different regional contexts.
This modules is appended to an pre-trained ResNet network, in our case a ResNet101.
In addition, ResNet was modified to use dilated convolutions to enlarge the field of view.
\\\textbf{Training}
All selected object detection networks are pre-trained on MS COCO \cite{978-3-319-10602-1_48}.
As configuration for each architecture we use the proposed settings from the framework \cite{8099834}.
Adjustments are made on the training settings. For all experiments, a value of 0.9 for the momentum optimizer \cite{qian1999momentum} is selected.
We apply cosine decay \cite{loshchilov2016sgdr} as learning rate strategy for the SSD meta architecture.
As parameters we select a learning rate and warmup learning rate of $3\mathrm{e}{-5}$ over $20,000$ steps.
The training for the R-FCN and Faster R-CNN is manually stopped if the performance on the validation set begins to saturate.
As learning rate strategy we reduce the learning rate by a factor of $10$ every $20,000$ steps.
The input dimensions for all networks change to a 3-channel RGB image with a fixed resolution of $640\times640$ pixels and batch size is set to 16.
For a better generalization of the fine-tuned model data augmentation methods are applied (\cite{krizhevsky2012imagenet,8202133,pmlr-v95-takahashi18a}).
We select random brightness, random contrast, random crop, random Gaussian noise and horizontal flip for all meta-architectures during the training.
For the semantic segmentation approach we use pre-trained ImageNet weights for PSPNet to fine-tune the architectures.
SegNet is trained from scratch without the usage of pre-trained weights.
As in our object detection setup we use the momentum optimizer with an momentum of 0.9 and a learning rate of 0.001.
Furthermore, the training uses a batch size of 4 and is done for 150 epochs.
The data augmentation methods random noise, horizontal flip and brightness are applied during training.

\section{Evaluation}
\label{sec:evaluation}

\setlength{\fboxsep}{0pt}
\begin{figure*}[ht!]
\captionsetup[subfigure]{labelformat=empty}
\centering
    \begin{sideways}
        \hskip 0.13\textwidth
        \small \(FES\)
    \end{sideways}
    \begin{subfigure}[b]{.25\linewidth}
        \caption{\(GT\)}
        \fbox{\includegraphics[width=\textwidth]{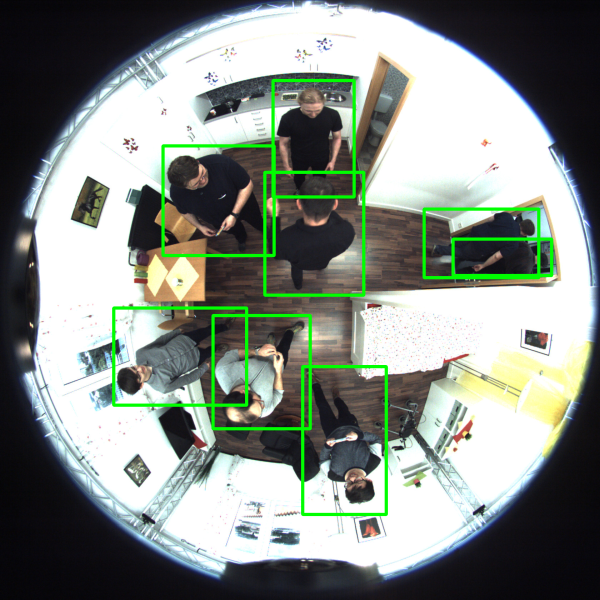}}
    \end{subfigure}
    \begin{subfigure}[b]{.25\linewidth}
        \caption{\(MS COCO\)}
        \fbox{\includegraphics[width=\textwidth]{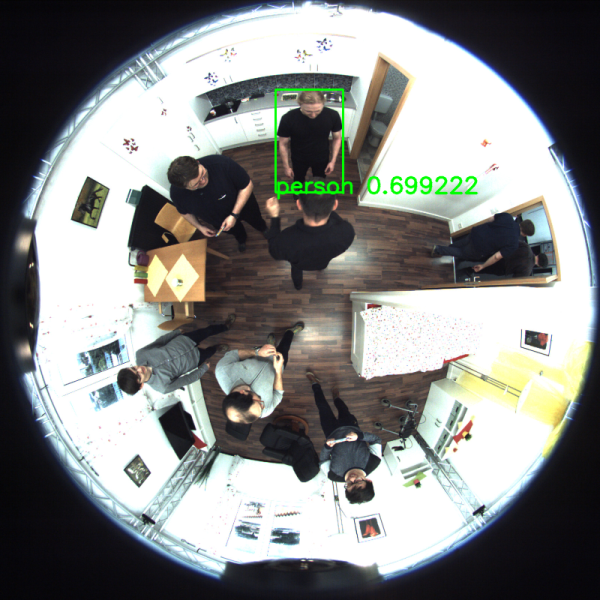}}
    \end{subfigure}
    \begin{subfigure}[b]{.25\linewidth}
        \caption{\(THEODORE\)}
        \fbox{\includegraphics[width=\textwidth]{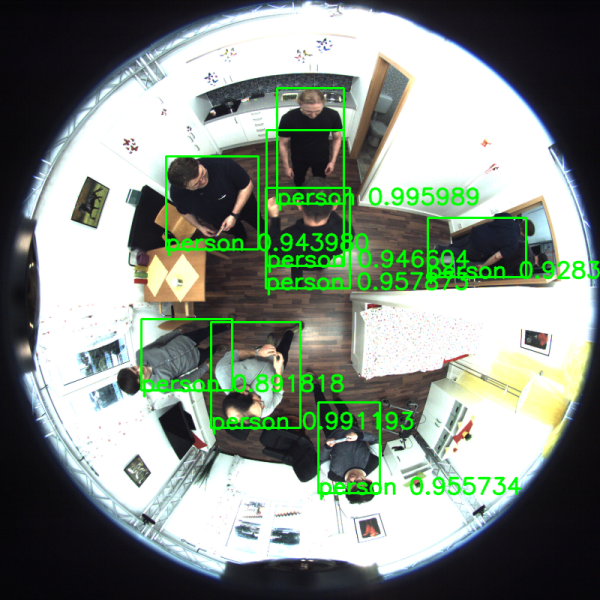}}
    \end{subfigure}

    \begin{sideways}
        \hskip 0.08\textwidth
        \small \(HDA\)
    \end{sideways}
    \begin{subfigure}[b]{.25\linewidth}
        \fbox{\includegraphics[width=\textwidth]{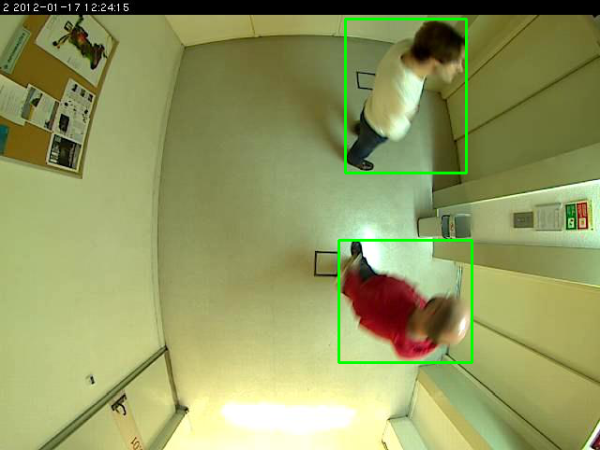}}
    \end{subfigure}
    \begin{subfigure}[b]{.25\linewidth}
        \fbox{\includegraphics[width=\textwidth]{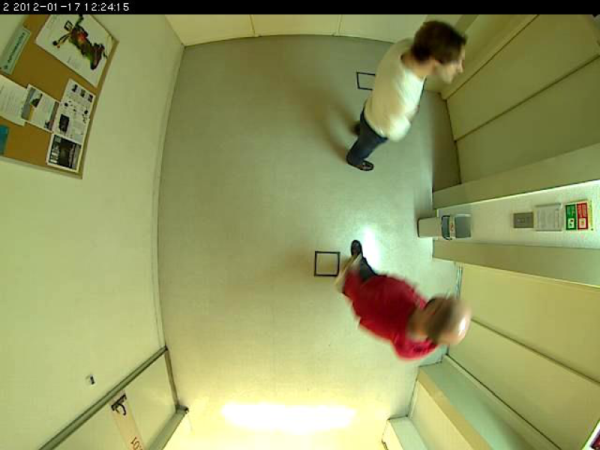}}
    \end{subfigure}
    \begin{subfigure}[b]{.25\linewidth}
        \fbox{\includegraphics[width=\textwidth]{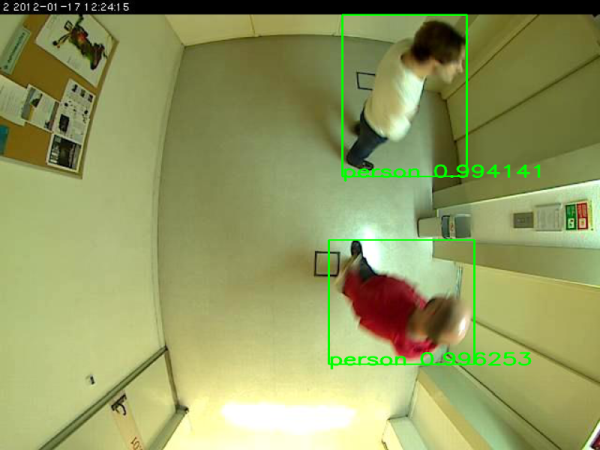}}
    \end{subfigure}

    \begin{sideways}
        \hskip 0.1\textwidth
        \small \(Bomni\)
    \end{sideways}
    \begin{subfigure}[b]{.25\linewidth}
        \fbox{\includegraphics[width=\textwidth]{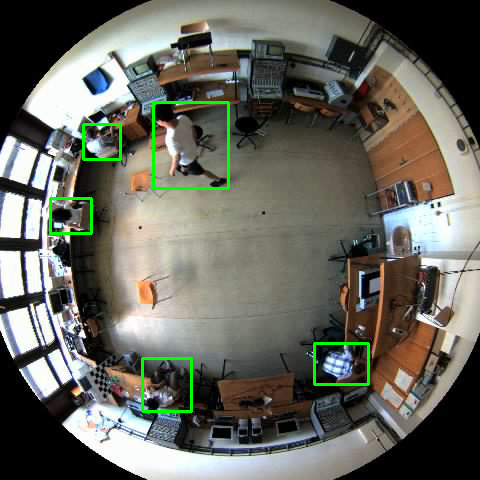}}
    \end{subfigure}
    \begin{subfigure}[b]{.25\linewidth}
        \fbox{\includegraphics[width=\textwidth]{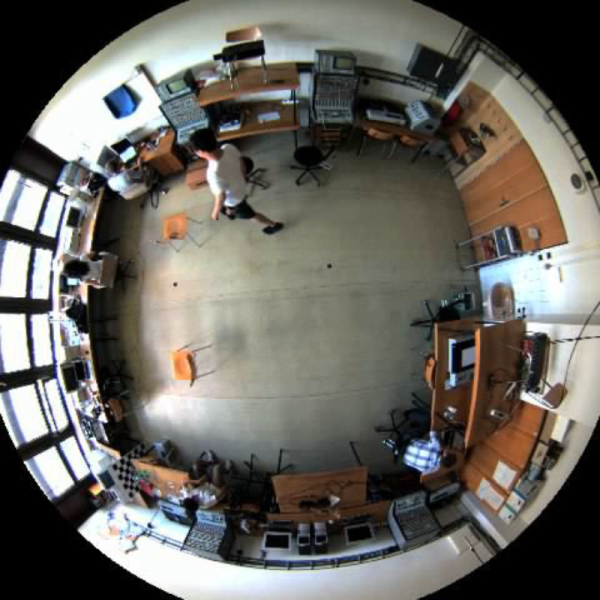}}
    \end{subfigure}
    \begin{subfigure}[b]{.25\linewidth}
        \fbox{\includegraphics[width=\textwidth]{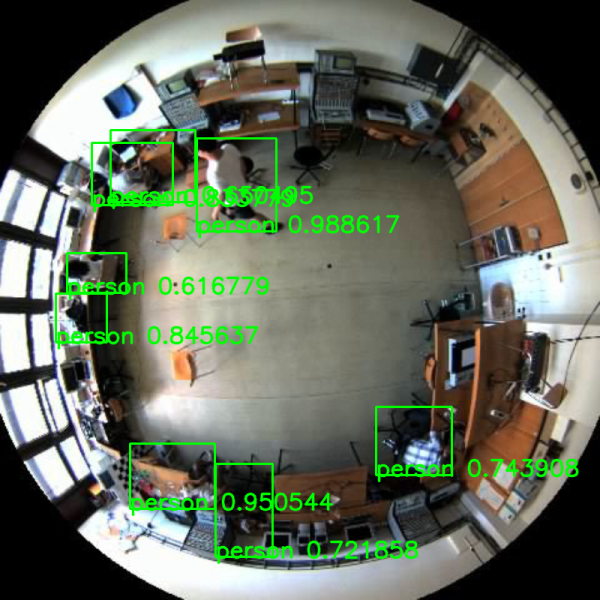}}
    \end{subfigure}

    \caption{Example of detection results on HDA, Bomni and FES dataset using SSD meta-architecture for person class.
    The first column contains the ground truth bounding boxes.
    In the second column the prediction results for the pre-trained CNN with MS COCO weights is shown.
    The last column indicates the detection boxes which we reach while fine-tuning on THEODORE.
    Statistics of the dataset and the evaluation through object detection is provided in the supplementary material.
    \label{fig:detection_results}}
\end{figure*}

\setlength{\fboxsep}{0pt}
\begin{figure*}[ht!]
\captionsetup[subfigure]{labelformat=empty}
\centering
    \begin{sideways}
        \hskip 0.05\textwidth
        \small \(Object\ Detection\)
    \end{sideways}
    \begin{subfigure}[b]{.2\linewidth}
        \fbox{\includegraphics[width=\textwidth]{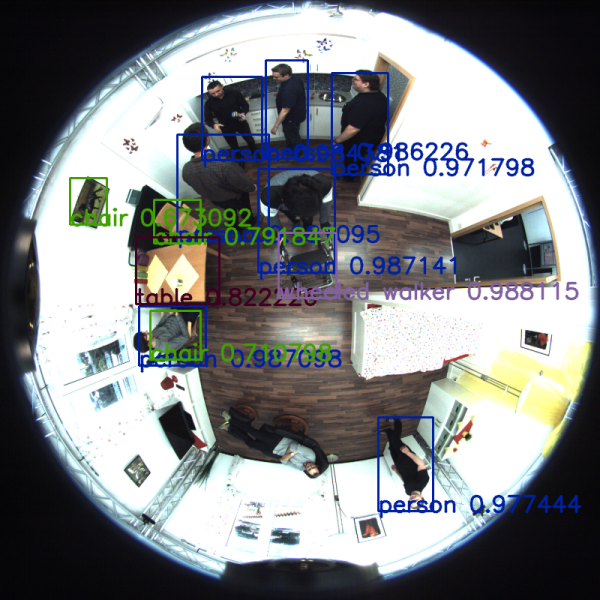}}
    \end{subfigure}
    \begin{subfigure}[b]{.2\linewidth}
        \fbox{\includegraphics[width=\textwidth]{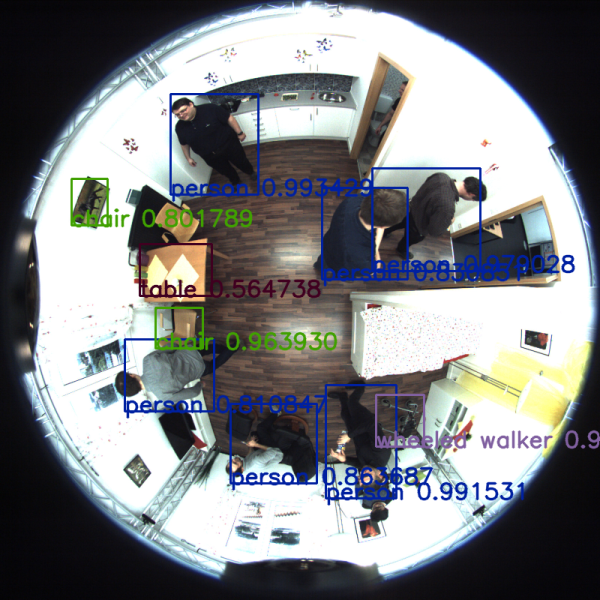}}
    \end{subfigure}
    \begin{subfigure}[b]{.2\linewidth}
        \fbox{\includegraphics[width=\textwidth]{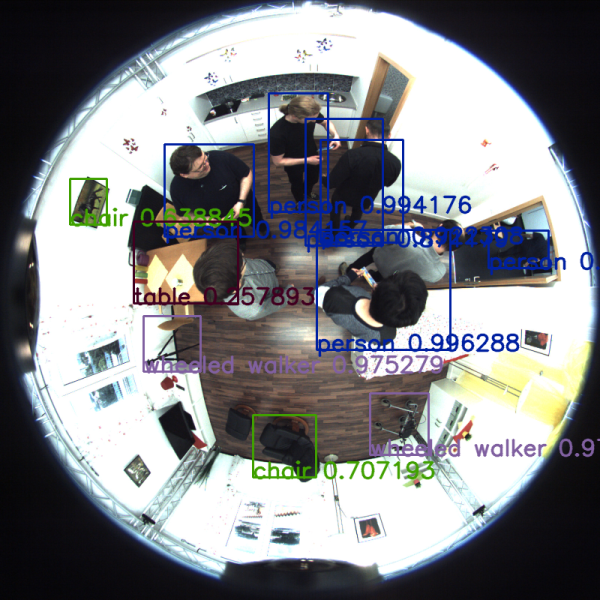}}
    \end{subfigure}
    \begin{subfigure}[b]{.2\linewidth}
        \fbox{\includegraphics[width=\textwidth]{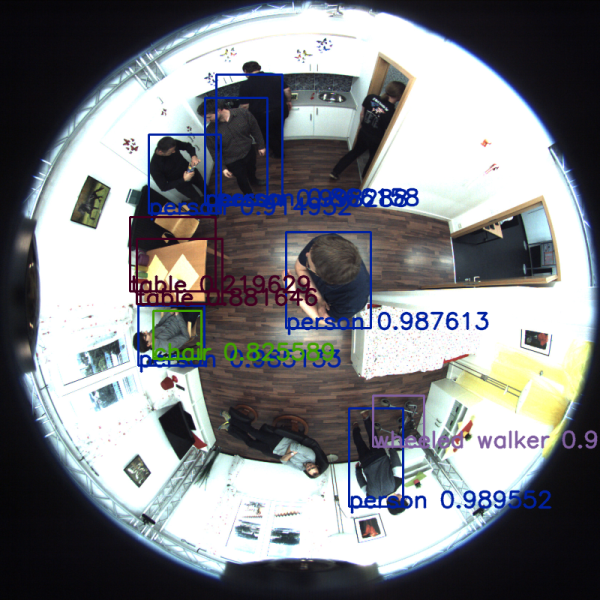}}
    \end{subfigure}

    \begin{sideways}
        \hskip 0.05\textwidth
        \small \(Segmentation\)
    \end{sideways}
    \begin{subfigure}[b]{.2\linewidth}
        \includegraphics[width=\textwidth]{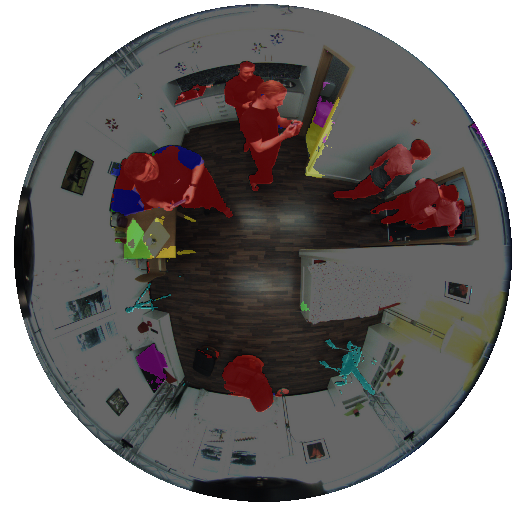}
    \end{subfigure}
    \begin{subfigure}[b]{.2\linewidth}
        \includegraphics[width=\textwidth]{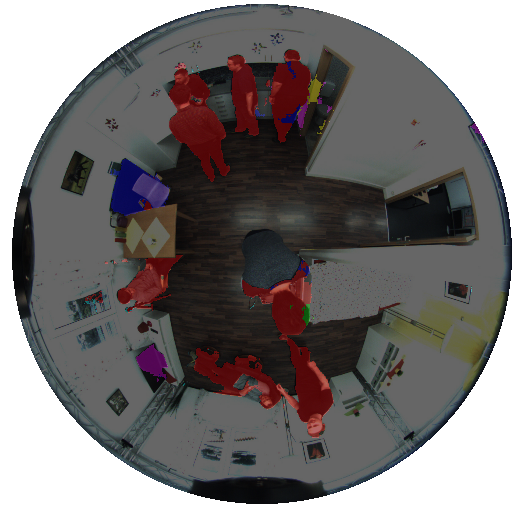}
    \end{subfigure}
    \begin{subfigure}[b]{.2\linewidth}
        \includegraphics[width=\textwidth]{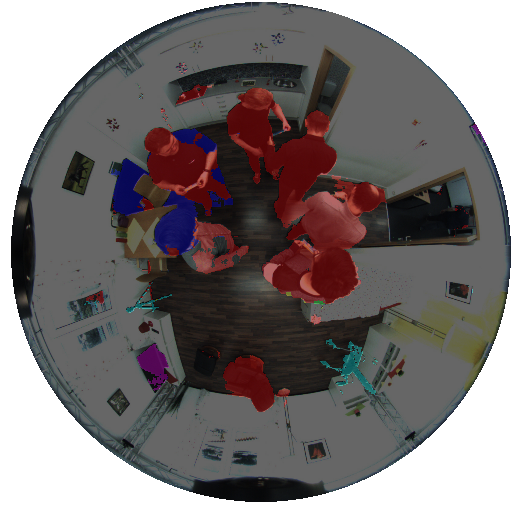}
    \end{subfigure}
    \begin{subfigure}[b]{.2\linewidth}
        \includegraphics[width=\textwidth]{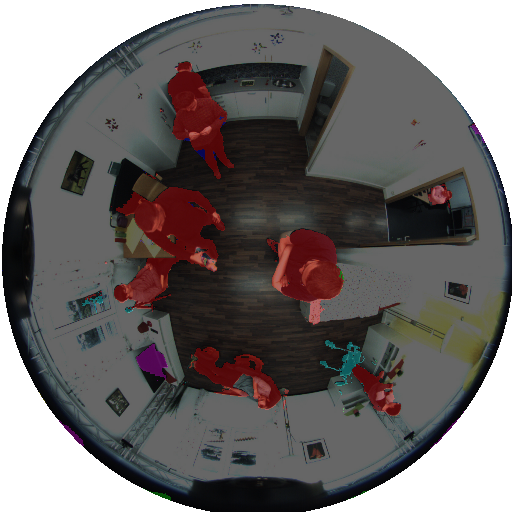}
    \end{subfigure}
    \begin{subfigure}[b]{.8\linewidth}
        \includegraphics[width=\textwidth]{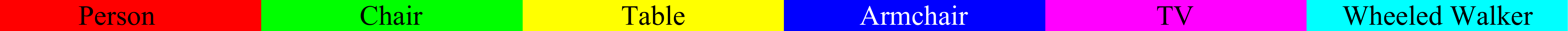}
    \end{subfigure}
    \caption{FES evaluation samples of object detection and segmentation architectures trained on THEODORE. The first row shows detection results from SSD, the second row the segmentation results of SegNet. SSD is trained on MS COCO and fine-tuned on THEODORE, while SegNet is trained from scratch on THEODORE.
    \label{fig:segmentation_results}}
\end{figure*}

With fine-tuning of CNNs with THEODORE we evaluate on labeled real world images by meta-architectures for object detection and segmentation as described in \autoref{sec:approach}.
We choose publicly available real world datasets such as High-Definition Analytics (HDA) \cite{nambiar2014multi} and Bo\u{g}azi\c{c}i University Multi-Omnidirectional (Bomni) \cite{demiroz2012feature} for object detection and an own annotated dataset Fisheye Evaluation Suite (FES) for semantic segmentation (see \autoref{sec:fisheyesuite}) to validate the meta-architectures fine-tuned on THEODORE.

As metric for evaluation the average precision (AP) \cite{everingham_pascal_nodate} per class and mean average precision (mAP) is reported for all classes.
Detections will be judged to be true positive, if the intersection over union (IoU) between the detected and ground truth bounding box is at least 0.5.
Our evaluation results shows exemplary bounding box detections in the first row of \autoref{fig:segmentation_results}.
For the evaluation on semantic segmentation we choose the mean intersection over union (mIoU) for the classes \textit{armchair}, \textit{chair}, \textit{person}, \textit{table}, \textit{tv} and \textit{wheeled walker}.
Exemplary results for semantic segmentation can be found in the second row of \autoref{fig:segmentation_results}.

\subsection{Number of images}
\begin{figure}[h!]
	\begin{center}
		\includegraphics[width=0.9\linewidth]{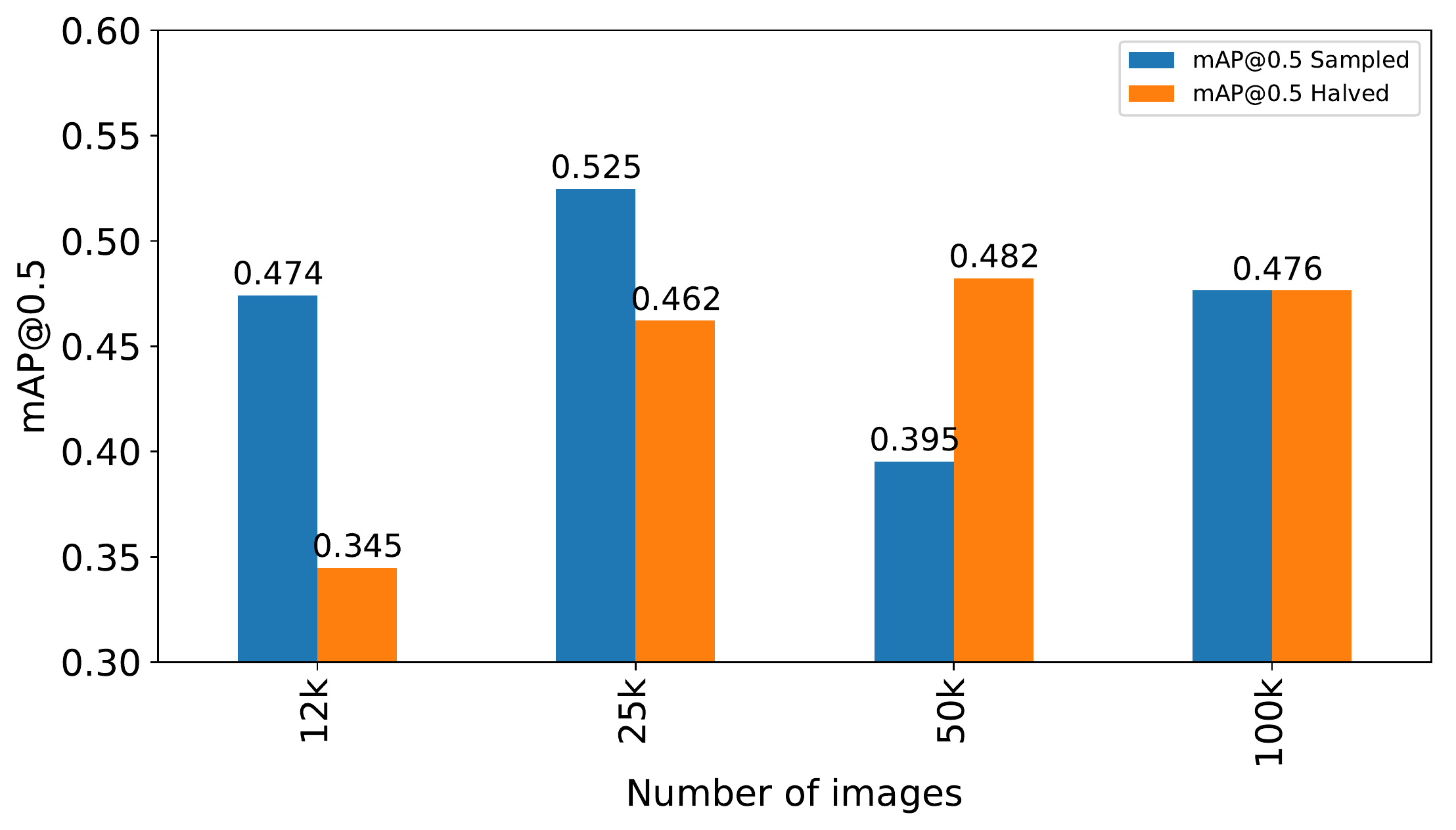}
    \end{center}
    \caption{Number of images in the training set with corresponding mAP@0.5 trained on the SSD meta-architecture. Sampled: constant number of textures with variable number of images; halved: halved number of images and halved number of textures \label{fig:ap_images_ratio}}
\end{figure}

In order to evaluate the amount of images that are relevant for THEODORE, we measured the mAP for the SSD meta-architecture validated on FES and summarize the results in \autoref{fig:ap_images_ratio}.
From the generated images we choose 12k, 25k, 50k and 100k images and train the SSD for $20,000$ steps each.
With the parameters described in \autoref{sec:dataset} the 100k images contain $500$ differently textured scenes in the training set.

In general, we carried out two approaches to reduce the number of images.
First, we sub-sample the 100k images to keep the number of differently textured scenes constant.
Second, the number of images is halved and consequently the number of textured scenes.
We observe that a increasing number of images not necessarily leads to better results.
The sub-sampled approach with 25k images results to the highest mAP.
This experiment shows that the number of scene variations have a higher impact on mAP than the absolute number of images for training.
For further experiments we use a subset of THEODORE with 25k images and 500 scenes.

\subsection{Object Detection}
\label{sec:fisheyesuite}
In this section we describe the evaluation of THEODORE on three real-world datasets, the HDA dataset, Bomni and FES.
The HDA and Bomni dataset only contains labels for the person class, so the evaluation was done on person class, as long as there are no other classes in the datasets publicly available.
The evaluation on our own dataset (FES) was done on six classes.

% \todo{Merge HDA, Bomni and DST into one section where we describe the AP for the Person class. Reason for that is that HDA and Bomni only contains labeled Persons.}
\noindent
\textbf{Validation on HDA} The HDA dataset \cite{nambiar2014multi} contains images captured with multiple cameras.
The dataset was created for the research on high-definition surveillance.
For our evaluation we use the 1388 labeled images from \textit{Cam 02}.
These images, with a resolution of $640 \times 480$ pixel were captured at 5\,Hz from the top-view position with a full 140\,\degree \ field of view.
The images of the HDA dataset are barrel distorted, which makes them more comparable to omnidirectional images.

\noindent
\textbf{Validation on Bomni}
Bomni Video Tracking Database contains video frames with a resolution of $640 \times 480$ pixel from an omnidirectional camera in a single room.
The dataset was created in the context of human tracking and action recognition. For our evaluation we use all frames from top-view cameras of scenario \#1 and crop them to a resolution of $480 \times 480$ pixel to remove most of the black borders.

\noindent
\textbf{Validation on FES}
To the best of our knowledge FES is the first dataset with real world fisheye top-view images. The dataset contains of 301 images and six class labels (\textit{person}, \textit{armchair}, \textit{chair}, \textit{table}, \textit{tv} and \textit{wheeled walker}) which were annotated manually.
All images have a resolution of $1680 \times 1680$ pixel with overlapping persons.
The images of the dataset, segmentation masks and bounding boxes are available at \url{https://www.tu-chemnitz.de/etit/dst/forschung/comp_vision/fes}.

\begin{table}[h]
\caption{\label{tab:od_person_eval} Quantitative evaluation of THEODORE for person class based on AP@0.5}
\begin{adjustbox}{width=1\columnwidth}
\begin{tabular}{ccccccc}
\multicolumn{1}{l||}{Person AP@0.5}                                                    & \multicolumn{3}{c||}{MS COCO} & \multicolumn{3}{c}{MS COCO + THEODORE} \\ \hline
\multicolumn{1}{l||}{}                                                       & HDA     & Bomni  & \multicolumn{1}{c||}{DST}    & HDA        & Bomni      & DST        \\ \hline
\multicolumn{1}{l||}{SSD}                                                    & 0.586   & 0.052  & \multicolumn{1}{c||}{0.484}  & \textBF{0.802}      & 0.579      & \textBF{0.904}      \\
\multicolumn{1}{l||}{R-FCN}                                                  & 0.303   & 0.069  & \multicolumn{1}{c||}{0.525}  & 0.694      & 0.675      & 0.849      \\
\multicolumn{1}{l||}{Faster R-CNN} & 0.627   & 0.067  & \multicolumn{1}{c||}{0.630}  & 0.704      & \textBF{0.740}      & 0.873      \\
\hline
\end{tabular}
\end{adjustbox}
\end{table}

For the evaluation of THEODORE we report the AP for person class for the HDA, Bomni and FES datasets in \autoref{tab:od_person_eval}.
As baseline, we choose MS COCO in the left three columns, while the right three colums indicate the APs with fine-tuning on THEODORE.
We achieve in all three meta-architectures for person class a significant improvement with THEODORE with respect to the baseline.

\begin{table}[h]
\caption{\label{tab:od_allclasses_eval} Quantitative evaluation of THEODORE for all classes based on mAP@0.5}
\begin{adjustbox}{width=1\columnwidth}
    \begin{tabular}{lccccccc}
    Class AP@0.5                                                    & Armchair & Chair & Person & Table & TV    & \multicolumn{1}{c||}{\begin{tabular}[c]{@{}c@{}}Wheeled\\ Walker\end{tabular}} & \multicolumn{1}{l}{mAP}   \\ \hline
    SSD                                                    & 0.021    & \textBF{0.231} & \textBF{0.904}  & 0.824 & 0.545 & \multicolumn{1}{c||}{0.623}                                                    & \multicolumn{1}{c}{0.525} \\
    R-FCN                                                  & \textBF{0.262}    & 0.039 & 0.849  & 0.859 & 0.000 & \multicolumn{1}{c||}{\textBF{0.640}}                                                    & \multicolumn{1}{c}{0.441} \\
    Faster R-CNN                                           & 0.148    & 0.141 & 0.873  & \textBF{0.980} & \textBF{0.943} & \multicolumn{1}{c||}{0.596}                                                    & \multicolumn{1}{c}{\textBF{0.613}} \\
    \hline
    \end{tabular}
\end{adjustbox}
\end{table}

The mAP of experiments on FES with six classes are shown in \autoref{tab:od_allclasses_eval}.
With 0.613 the highest mAP is reached with the Faster R-CNN.
The per-class winners are highlighted bold in \autoref{tab:od_allclasses_eval}.
The classes person and table have the highest APs which can explained through a good representation in the training data, i.e. various viewing angle and texture.
Improvements needs to be done in the classes armchair, chair and TV.
The low AP values can have different reasons.
First, the objects in the training data have too less variations in terms of illumination, texture and viewing angle.
Second, the training data doesn't fit well to the test data, which ends up with the creation of a more generalized model for these classes.
Another effect we observed through the evaluation is the non-detection of the class TV with R-FCN.
For this we suspect a too high shrinking of the image as input for the net, so the filter sizes are to big for the whole image to detect small objects with the R-FCN.

\subsection{Semantic Segmentation}
Beside object detection we show that THEODORE is eglible for training segmentation networks.
Due to the lack of publicy available top-view fisheye label masks for evaluation of THEODORE we annotate own data, namely FES.
The report of the class IoU and mIoU on two state of the art architectures for segmentation, SegNet and PSPNet, is shown in \autoref{tab:seg_eval}.
\begin{table}[h]
\caption{\label{tab:seg_eval} Quantitative evaluation of THEODORE by fine-tuning CNN meta-architectures for semantic segmentation.}
\begin{adjustbox}{width=1\columnwidth}
    \begin{tabular}{lccccccc}
    Class IoU                                                    & Armchair & Chair & Person & Table & TV    & \multicolumn{1}{c||}{\begin{tabular}[c]{@{}c@{}}Wheeled\\ Walker\end{tabular}} & \multicolumn{1}{l}{mIoU}   \\ \hline
    SegNet                                        & 0.009    & 0.016 & \textBF{0.674}  & 0.012 & 0.53 & \multicolumn{1}{c||}{0.33}                                                    & \multicolumn{1}{c}{\textBF{0.359}} \\
    PSPNet                                        & 0.005    & 0.023 & 0.434  & 0.003 & 0.195 & \multicolumn{1}{c||}{0.034}                                                    & \multicolumn{1}{c}{0.229} \\
    \hline
    \end{tabular}
\end{adjustbox}
\end{table}

In \autoref{tab:seg_eval} we evaluate THEODORE by fine-tuned SegNet and PSPNet on the FES.
We observe class IoUs of 0.67 for person and 0.53 for TV.
The mIoU lies at 0.36 for the SegNet and 0.23 for the PSPNet.
Both segmentation architectures are realtively good in the class person, while classes like chair, armchair and table needs further investigations.
We belief that the texture of the synthetically generated furniture is different from the real-world furniture texture.

\section{Conclusion}
\label{sec:conclusion}
In this paper we introduce \textit{THEODORE - a synTHEtic tOp-view inDoOR scEnes} dataset with omnidirectional images.
This dataset contains 100,000 rendered images of diversified indoor environments, segmentation and instance masks for 16 classes and bounding boxes for the person class.
Additionally, we have shown that the usage of synthetically generated images could compensate the lack of real omnidirectional images during training of CNNs.
We have addressed the task of object detection and semantic segmentation for evaluating the performance of state-of-the-art CNNs trained on THEODORE.
The evaluation process of our dataset works as follows: the training baseline is MS COCO, which contains front-views of perspective images.
We fine-tune three meta-architectures for object detection, namely SSD, R-FCN and Faster R-CNN for the person class on THEODORE.
In addition we train two meta-architectures for semantic segmenation, the SegNet and PSPNet for six classes in an indoor environment.
Both object detectors and segmentation approaches were evaluated on our own annotated fisheye evaluation suite dataset (FES), that contains segmentation and object detection ground truth for six classes.
With this we have shown the adaptation of the front-view to the top-view by fine-tuning CNNs with our generated data.
While labels for fixed objects are not available in public real world databases, we use six classes for evaluation of THEODORE, which leads to significant improvement of the AP and mIoU over the baselines in all tested meta-architectures.

Future research will address the balancing of the classes of THEODORE.
While the FES evaluation dataset only contains one scenario, we plan to add more real world indoor scenes.
Beyond the segmentation and detection masks we intend to create omnidirectional depth, skeletons and optical flow ground truth from rendered scenes.

{\small
    \bibliographystyle{ieee}
    \bibliography{paper.bib}
}

\end{document}